\documentclass[letterpaper, 10 pt, conference]{ieeeconf}  

\IEEEoverridecommandlockouts                             

\usepackage{ifpdf}
\usepackage{cite}
\usepackage{graphicx}
\usepackage{amsmath, amssymb}
\usepackage[linesnumbered, ruled]{algorithm2e}
\usepackage{array}
\usepackage{mdwmath}
\usepackage{mdwtab}
\usepackage{eqparbox}
\usepackage{subcaption}
\usepackage{stfloats}
\usepackage{url}
\usepackage{multirow}
\usepackage{multicol}
\usepackage{mathtools}
\usepackage{color}
\usepackage{flushend}

\newcommand{\norm}[1]{{\left\lVert#1\right\rVert}}

\title{\LARGE \bf
The Globally Optimal Reparameterization Algorithm: an Alternative to Fast Dynamic Time Warping for Action Recognition in Video Sequences
}

\author{Thomas W. Mitchel*, Sipu Ruan*, Yixin Gao, Gregory S. Chirikjian
\thanks{*Equally contributing authors} \\
\thanks{Thomas W. Mitchel, Sipu Ruan and Gregory S. Chirikjian are with the Laboratory of Computational Sensing and Robotics, Johns Hopkins University, Baltimore, Maryland 21218 {\tt\small \{tmitchel, ruansp, gchirik1\}@jhu.edu}}
\thanks{Yixin Gao is with the Department of Computer Science, Johns Hopkins University, Baltimore, Maryland 21218}
}

\begin{document}
\maketitle
\thispagestyle{empty}
\pagestyle{empty}
\begin{abstract}
Signal alignment has become a popular problem in robotics due in part to its fundamental role in action recognition. Currently, the most successful algorithms for signal alignment are Dynamic Time Warping ({\it DTW}) and its variant 'Fast' Dynamic Time Warping ({\it FastDTW}). Here we introduce a new framework for signal alignment, namely the Globally Optimal Reparameterization Algorithm ({\it GORA}). We review the algorithm's mathematical foundation and provide a numerical verification of its theoretical basis. We compare the performance of {\it GORA} with that of the {\it DTW} and {\it FastDTW} algorithms, in terms of computational efficiency and accuracy in matching signals. Our results show a significant improvement in both speed and accuracy over the {\it DTW} and {\it FastDTW} algorithms and suggest that {\it GORA} has the potential to provide a highly effective framework for signal alignment and action recognition. 

\end{abstract}

\section{INTRODUCTION} \label{intro}
With the recent emergence of new machine learning techniques, there has been an increasing interest in robotic action recognition.  The foundation of action recognition lies in the problem of signal alignments, in the sense that prior to categorizing or identifying  sets of sequences, one must establish a method to temporally parameterize the sequences that enables standardized comparisons.

Currently the most successful techniques for signal alignments are based on the well-known method of \textit{Dynamic Time Warping} ({\bf DTW}) \cite{sakoe1978dynamic}, which matches two time series with a monotonically increasing optimal warping path satisfying boundary conditions. Since its introduction almost 40 years ago, DTW has been applied to a variety of fields including speech recognition \cite{shaikh2017recognition}, action recognition \cite{sempena2011human}, data mining \cite{keogh2000scaling}, and motion perception \cite{zhu2017robust}. Due to DTW's $O(T^{2})$ time complexity, many variants have been introduced with the goal of striking a balance between accuracy and computational efficiency \cite{keogh2001derivative,salvador2007toward,keogh2005exact}, the most widely-used of these being the \textit{Fast Dynamic Time Warping} ({\bf FastDTW}) algorithm, which achieves a time complexity of $O(T)$ \cite{salvador2007toward}. One of the most recent developments in the DTW family has been the introduction of 
\textit{Generalized Time Warping} \cite{zhou2012generalized}, which aligns multiple multi-modal sequences with linear time complexity.

Recently, a novel, alternative mathematical framework for signal alignment has been proposed, in which signals are reparameterized to a \textit{universal standard timescale} ({\bf UST}) using principles of variational calculus \cite{chirikjian2017signal}. The goal of this paper is to introduce an efficient numerical algorithm for signal alignment based on this framework, which we will henceforth refer to as the \textit{Globally Optimal Reparameterization Algorithm} ({\bf GORA}), and to provide an initial numerical validation of this approach.  

Given two or more time-evolving signal sequences, GORA temporally reparameterizes each to a UST that allows for pairwise comparison at each instance in time. Re parameterizations are found using variational calculus to produce mappings to new temporal variables that globally minimize the amount of change in the sequences, representing a new approach to the problem of signal alignment. The major advantages of this approach are: \\
\textit{
1. It achieves linear time complexity of $O(T)$, where T is the number of time instances in a signal;\\
2. It can simultaneously reparameterize multiple signal sequences to a universal time scale; and\\ 
3. It can potentially be built-upon to allow for the effects of nuisance parameters such as noise or motion artifacts to be minimized or eliminated \cite{chirikjian2017signal}. }

The remainder of the paper is organized as follows.  First, we review GORA's mathematical foundations, which were first described in \cite{chirikjian2017signal}, and define and introduce GORA itself. We then discuss the settings for its application to signals in the form of both real trajectories and video sequences. This is followed by a numerical verification of GORA's ability to find globally optimal temporal reparameterization of a given signal. We then provide an initial verification of the algorithm by comparing its performance, in terms of both computational efficiency and accuracy in matching signals, relative to DTW methods. Using both synthetic and real datasets, our results show a significant improvement in both speed and accuracy over the DTW and FastDTW algorithms. We conclude with a short discussion on the computational significance of the differences between GORA and DTW methods, in addition to the authors' plans for the continued development of the GORA framework. 

\section{PROBLEM STATEMENT}
Without loss of generality, consider any kind of temporally evolving signal, $X(t)$, as a mapping from the unit interval to the space $S$, i.e. $X:[0, 1] \, \rightarrow \, S$,  on which that particular type of signal evolves. Defining a metric $d$ on $S$, $(S, d)$ becomes a metric space. In general, given any two signals $X_{1}(t)$, $X_{2}(t)$ it is likely that $$ \int_0^1 d(X_{1}(t), X_{2}(t)) \ dt \gg 1, $$ even if it is suspected that both signals portray similar dynamic phenomena, a major reason being that each signal could have a different temporal parameterization on the the unit interval. 

The GORA algorithm is based on the notion that the temporal misalignment between two arbitrary signals can be compensated by reparameterizing each to a UST. In other words, assuming that nuisance parameters or motion artifacts, such as variations in perspective, are not significantly affecting the signals, if we can find two strictly monotonically increasing functions, $\tau_{1}^*, \tau_{2}^* \in C^{1}[0, 1]$ such that  $$ \int_0^1 d(X_{1}(\tau_{1}^*(t)), X_{2}(\tau_{2}^*(t))) \ dt \ll 1, $$ then we can say that $X_1(t)$ and $X_2(t)$ are fundamentally the same. 

Let ${\cal T}$ represent the set of all such $C^{1}$ monotonically increasing functions on the unit interval. Denoting $\circ$ as the operation of composition of functions, namely $$ (\tau_1 \circ \tau_2) (t) = \tau_1(\tau_2(t)) \quad \forall \ \tau_1, \tau_2 \in \cal{T},$$ $({\cal T}, \circ)$ forms a group, which we refer to as the \textit{temporal reparameterization group} ({\bf TRG}).  For a given signal, $X(t) \in S$, one can use the succeeding variational calculus formulation to find a globally optimal $\tau^{*} \in {\cal T}$ such that $X(\tau^{*}(t))$ is the UST parameterization of $X$, reducing a search for this mapping from $S$ to the quotient space $S/{\cal T}$. 

\section{MATHEMATICAL FORMULATIONS}
Suppose one wants to find a function, ${\bf x}(t)$, that extremizes
a functional of the form
\begin{equation}
J = \int_{0}^{1} f\left({\bf x},\dot{\bf x},t\right)\,dt
\label{eq:cost_functional}
\end{equation}
where $ \dot{\bf x} = d{\bf x}/dt$.
This type of problem can be addressed by the application of Calculus of Variations, and the desired ${\bf x}(t)$ is the solution to the Euler-Lagrange equations:
\begin{equation}
\frac{\partial f}{\partial {\bf x}} -
\frac{d}{dt} \left(\frac{\partial f}{\partial \dot{\bf x}} \right) = {\bf 0}
\label{eulerlag}
\end{equation}
In general, there are no guarantees that the solution to the preceding equations will be globally optimal, however, in certain situations (including optimal temporal reparameterization), the structure of the function $f(\cdot)$ will guarantee that the solution generated by the equations is in fact a globally optimal solution. The following theorem is an example of one such case. 

\subsection{Theorem and Proof of Global Optimality}
\noindent
{\bf THEOREM 1}:
{\it If $x: [0, 1] \, \rightarrow \, \mathbb{R}$ and the integrand in the cost functional \eqref{eq:cost_functional} is of the form
\begin{equation}
f(x,\dot{x}) = \dot{x}^2 \mathfrak{g}(x)
\label{optcost}
\end{equation}
where $\mathfrak{g}(x):\mathbb{R} \, \rightarrow \, \mathbb{R}_{>0}$ is $C^1$, then the solution generated by \eqref{eulerlag} subject to the boundary conditions $x(0) = 0$ and $x(1) =1$ is globally minimal}.
\\
\\
The proof of this theorem was first demonstrated in \cite{chirikjian2017signal}, however we choose to re-demonstrate it here as it illuminates the fundamental structure of GORA.
\\
\\
\begin{proof} Evaluating (\ref{eulerlag}) with (\ref{optcost}) gives
\begin{equation}
 2 \ddot{x} \mathfrak{g} + \dot{x}^2 \frac{\partial \mathfrak{g}}{\partial x} = 0.
 \label{origode}
\end{equation}
Multiplying both sides by $\dot{x}$ and integrating yields the
exact differential
$$ \frac{d}{dt}(\dot{x}^2 \mathfrak{g}) = 0. $$
Integrating both sides with respect to $t$ and isolating $\dot{x}$ yields
$$ \dot{x} = c\, \mathfrak{g}^{-\frac{1}{2}}(x) $$
where $c$ is the arbitrary constant of integration. With the
boundary conditions $x(0) =0$ and $x(1) =1$, we can then
write
$$ F(x^*) \doteq \frac{1}{c} \int_{0}^{x^*} \mathfrak{g}^{\frac{1}{2}}(\sigma)\,d\sigma = t, $$
where
$$ c = \int_{0}^{1} \mathfrak{g}^{\frac{1}{2}}(\sigma)\,d\sigma. $$
The notation $x^*$ indicates that this is the unique solution obtained from the
Euler-Lagrange equations that satisfies the boundary conditions.

The function  $F(x^*) = t$ can be inverted ($F$ is monotonically increasing since
$\mathfrak{g}(x) > 0$) to yield $x^* = F^{-1}(t).$

To see that this solution is globally optimal, substitute
\begin{equation}
\dot{x}^* = \mathfrak{g}^{-\frac{1}{2}}(x^*) \int_{0}^{1} \mathfrak{g}^{\frac{1}{2}}(\sigma)\,d\sigma
\label{soln}
\end{equation}
into the cost functional
$$ J(y) = \int_{0}^{1} \mathfrak{g}(y) \dot{y}^2 dx $$
where $y(t)$ is \underline{any} function in ${\cal T}$.
Then
$$ J(x^*) = \left(\int_{0}^{1} \mathfrak{g}^{\frac{1}{2}}(x^*)\,dx^* \right)^2
= \left(\int_{0}^{1} \mathfrak{g}^{\frac{1}{2}}(y)\,dy \right)^2, $$
where the second equality is simply a change of name of the dummy variable of integration.
Furthermore, since $x^*$ and $y$ are both functions of time, we can change the domain of integration as
$$  J(x^*) = \left(\int_{0}^{1} \mathfrak{g}^{\frac{1}{2}}(y(t)) \,\dot{y}\, dt \right)^2. $$
Since in general, from the Cauchy-Schwarz inequality,
$$ \left(\int_{0}^{1} f(t)\,dt \right)^2 \,\leq \int_{0}^{1} [f(t)]^2 dt, $$
we see that by letting $f(t) = \mathfrak{g}^{\frac{1}{2}}(y) \dot{y}$ that
$$ \left(\int_{0}^{1} \mathfrak{g}^{\frac{1}{2}}(y(t)) \,\dot{y}\, dt \right)^2 \,\leq\,
\int_{0}^{1} \mathfrak{g}(y) (\dot{y})^2 dt $$
and hence
$$ J(x^*) \leq J(y) $$
where $x^*(t)$ is the solution generated by the Euler-Lagrange
equation and $y(t)$ is any function in ${\cal T}$. Therefore $x^*(t)$ is a globally minimal solution. 
\end{proof} 

\subsection{The Globally Optimal Reparameterization Algorithm (GORA)} \label{algorithm}
In the context of signal alignment, the solution to the preceding variational problem provides a method for finding the UST parameterization of a given signal.  In particular, taking $x \in \cal{T}$ we have $x^* = \tau^*$,  subject to the definition of $\mathfrak{g}(x)$, which measures the rate of change of the given signal along the temporal axis. This is the backbone of Globally Optimal Reparameterization Algorithm (GORA), defined in Algorithm \ref{algo1}.

\begin{algorithm}
\SetKwInOut{Input}{Input}
\SetKwInOut{Output}{Output}

\caption{Globally Optimal Reparameterization Algorithm (GORA)}
\label{algo1}
\Input{Input signal $X(t)$; Initial temporal variable $t$}
\Output{UST reparameterization of signal $X^{*}(t)$}

Compute $\mathfrak{g}(t)$;\\
$c = $ {\it NumericalIntegration}($\mathfrak{g}^{\frac{1}{2}}(\sigma)$, $[0,1]$); \\
$F(\tau^*) = \frac{1}{c} \ {\it NumericalIntegration}(\mathfrak{g}^{\frac{1}{2}}(\sigma)$, $[0,\tau^*]$); \\
$\tau^*(t) = F^{-1}(t)$; \\
$X^{*}(t) = {\it Interpolation}(X(t), \tau^*(t))$;
\end{algorithm}
Given $\mathfrak{g}(t)$, calculating $\tau^*(t)$ is relatively straightforward and follows the first part of the proof of Theorem 1. For a given signal, the function $\mathfrak{g}(t)$ should be defined analogous to the squared magnitude of the temporal derivative of the signal. For example, in the case of a video signal, an appropriate definition of $\mathfrak{g}(t)$ could be based on the temporal derivative of the matrix of pixel values representing each frame. 

It should be noted that steps 2-3 in GORA can be performed simultaneously. Additionally, the method of interpolation through which $X^*(t)$ is recovered from $X(t)$ and $\tau^*(t)$ in step 5 should be chosen based on the properties of the input signal.

\section{Numerical Verification Settings}
In this paper, we provide a validation of GORA using discretized signals, in the form of both synthetically generated trajectories in $\mathbb{R}^3$ and video sequences from the Weizmann Action Recognition Classification Database \cite{ActionsAsSpaceTimeShapes_iccv05,ActionsAsSpaceTimeShapes_pami07}. The following section describes our experimental regime and results.

\subsection{Signal structure} The version of GORA implemented in our experiments is designed for signals in the form of real trajectories. Through vectorization, each frame of a video sequence can be represented as an $n \times 1$ array of pixel values where $n=\textit{width} \times \textit{height}$. As such, any video sequence can be described by a temporally-evolving curve in $\mathbb{Z}^n$. If we imagine video sequences as collections of discretized samplings of continuous phenomena at arbitrary time instances, any re-sampling at new time instances produces a curve in $\mathbb{R}^n$.

Additionally, it is important to note that for the sake of sampling consistency, we trimmed all video sequences in the Weizmann Database such that each trimmed video showed only a \textit{single instance of an action being performed}. For example, videos of a person  walking were trimmed to show only a single stride (two successive placements of the same foot) and videos of a person waving multiple times were trimmed to show only a single wave. 

\subsection{Formulation of $\mathfrak{g}(t)$}
For signals of the form $X(t) \in \mathbb{R}^n$, a natural choice for the definition of $\mathfrak{g}(t)$ consistent with~(\ref{optcost}) is 
\begin{equation}
\mathfrak{g}(t) = \left\| \frac{dX}{dt} \right\|^2, 
\end{equation}
where $\| \cdot \|$ denotes the Euclidean norm of a vector. In practice, we computed $ d X / d t$ using a high order finite difference method.

\subsection{Error metric}
Given two discretized signals $\{X_1\}, \{X_2\} \in \mathbb{R}^n$ we defined the distance or error between them as the average euclidean distance over all time instances, namely,
\begin{equation}
error_{\mathbb{R}^{n}} = \frac{1}{T}\sum_{i = 1}^{T} \norm{\{X_1\}_i - \{X_2\}_i}, \label{error_seq}
\end{equation}
where $T$ is the number of time instances.

\section{Verifications of Global Optimality}
\begin{figure*}[!t] 
\centering
\begin{subfigure}{.48\linewidth}
\includegraphics[width=\textwidth]{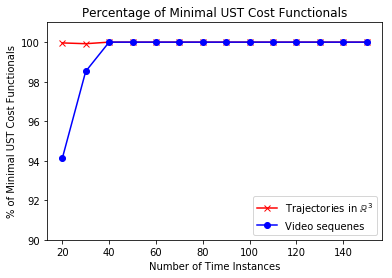}
\caption{Percentages of UST cost functional values lower than initial cost functional values for increasing numbers of time instances in input signals.}
\label{fig:opt_pctg}
\end{subfigure}
\vskip \baselineskip
\begin{subfigure}{.48\linewidth}
\includegraphics[width=\textwidth]{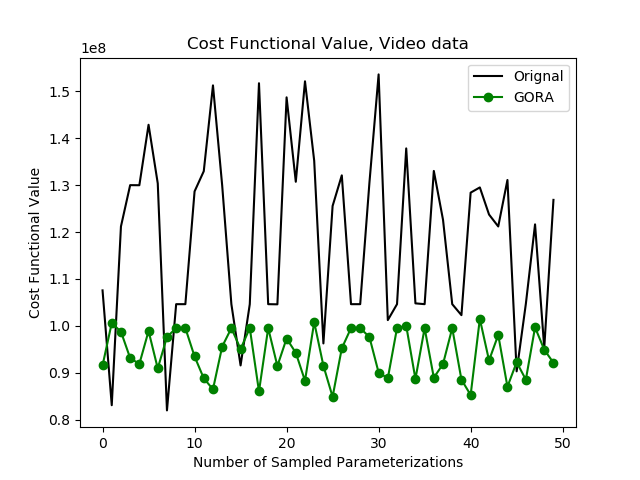}
\caption{The cost functional values for 50 different initial parameterizations of a template video sequence with 20 times instances and their corresponding UST reparameterizations.}
\label{fig:cost_video20}
\end{subfigure}
\hskip.5em
\begin{subfigure}{.48\linewidth}
\includegraphics[width=\textwidth]{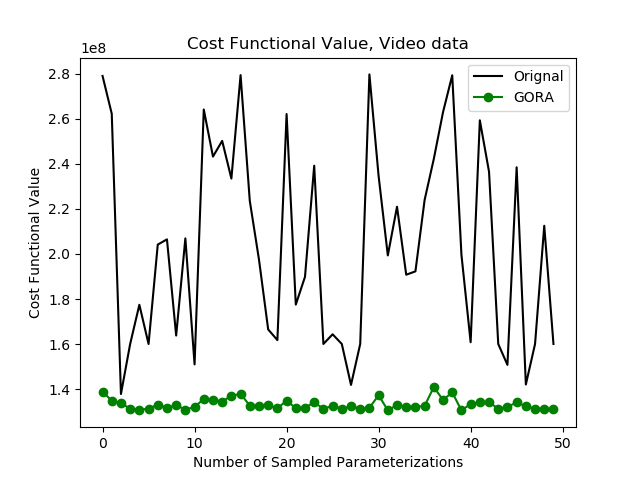}
\caption{The cost functional values for 50 different initial parameterizations of the same template video sequence with 100 times instances and their corresponding UST reparameterizations.}
\label{fig:cost_video100}
\end{subfigure}

\caption{Comparisons between the cost functional values of randomly parameterized input signals and their UST reparameterized counterparts found by GORA.}
\label{fig:cost}
\end{figure*}

One of the major advantages of GORA is its ability to reparameterize multiple signals to their corresponding USTs in parallel, which allows for pairwise comparison between signals. Here we seek to verify the global optimality of UST parameterizations computed by GORA using both synthetic curves in $\mathbb{R}^3$ and video sequences in the form of vectorized curves in $\mathbb{R}^n$.

In principle, for an arbitrary input signal its UST reparameterization found by GORA minimizes the integrand in~(\ref{eq:cost_functional}) with with respect to the cost function given by~(\ref{optcost}). Our experimental procedures for evaluation are summarized as follows: For a given number of time instances, we randomly selected 50 template signals. For each template signal, $X_0(t)$, we randomly generated 50 functions in the TRG and reparameterized $X_0(t)$ with respect to each to create 50 different input signals. We then used GORA to obtain the UST parameterization, $\tau^*(t)$, and recover the the UST reparameterized version of the signal, $X^*(t)$. We then computed the value of the cost functional with respect to $\tau^*(t)$ and $X^*(t)$ and compared this with the value of the cost functional computed using the input signal and original timescale.

The results of our global optimality experiments are displayed in Fig. \ref{fig:cost}.  Fig. \ref{fig:opt_pctg} shows the percentages of UST computed cost functional values lower than the cost functional values computed using the initial signals and initial timescales from 20 to 150 time instances. At each time instance, the percentages computed with respect to the 2500 (50 $\times$ 50) input and corresponding UST signal pairs generated using the experimental procedures detailed in the preceding paragraph. The red and blue lines show the results for synthetic trajectories in $\mathbb{R}^3$ and for vectorized video sequences, respectively.

As an example, Figs. \ref{fig:cost_video20} and \ref{fig:cost_video100} show the values of the cost functionals for fifty different pairs of randomly parameterized input signals and their corresponding UST reparameterizations found by GORA, with 20 and 100 time instances respectively. For both Figs \ref{fig:cost_video20} and \ref{fig:cost_video100}, all input signals were generated from a single template video sequence in the Weizmann Database. The black and green lines represent the cost functional values for the input signals and their UST reparameterizations, respectively.

The results indicate that in general, GORA does a remarkably good job of finding UST parameterizations that are globally minimal (or at least very close, depending on numerical precision) in the sense of~(\ref{eq:cost_functional}). GORA's failure to so consistently in the case of signals with low numbers of time instances (e.g. Fig. \ref{fig:cost_video20}) can likely be explained by its reliance on numerical differentiation of the input signal (in the computation of $\mathfrak{g}(t)$) and on numerical integration of $\mathfrak{g}(t)$, both of which become less accurate with larger temporal step sizes corresponding to lower numbers of time instances.

Additionally, this type of numerical evaluation between the values of the UST computed cost functionals and cost functionals computed using the initial sequences and timescales might provide a template for finding the lower bound of GORA's effectiveness with a given signal type. When the UST parameterization found by GORA is clearly not globally optimal, as is the case when the cost functional computed with respect to the input signal has a lower value than that computed using the UST reparameterization, it cannot be considered to be accurate. When using GORA for pairwise signal comparisons, failure to well-approximate UST parameterizations would likely lead to a greater degree of induced error. Depending on the properties of the input signals, chosen methods of derivation, integration, and interpolation, one might be able to probe for a lower bound on the number of time instances based on a desired accuracy threshold. 

\section{Algorithm Performance and Comparisons}
\label{Comparisons}
This section summarizes our comparisons between the performance of GORA and that of the DTW and FastDTW \cite{salvador2007toward} algorithms. Specifically, we evaluate the performance of each of the above algorithms in terms of both accuracy in matching signals and computational efficency. All comparisons are performed in Python 2.7 and the DTW and FastDTW implementations we used in our experiments were from the official Python packages. The experiments were performed on an Intel Core i7-7600U CPU @ 2.80GHz. 

\subsection{Comparison regime}
\begin{figure*}[!t]
\centering
\includegraphics[]{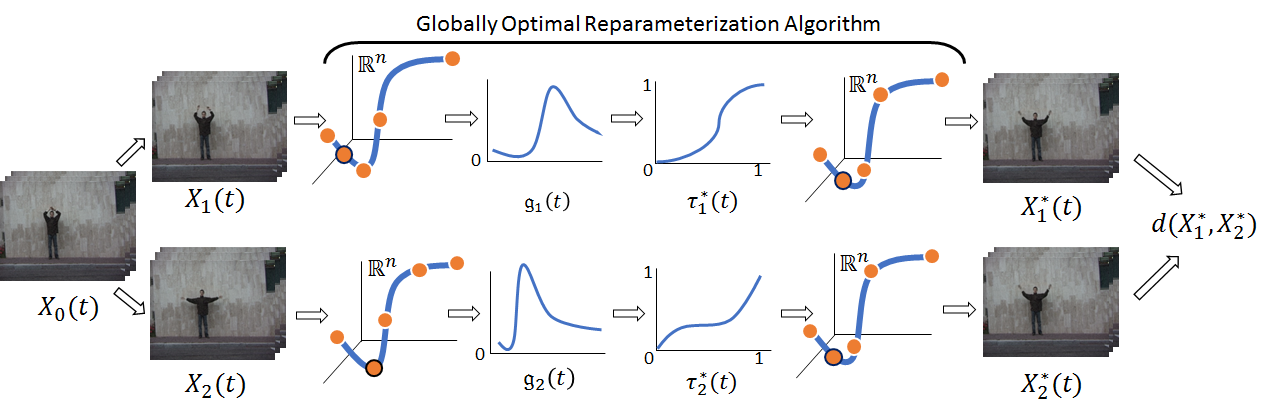}
\caption{The experimental work flow for evaluating GORA's computational efficiency and accuracy in matching signals: A template signal is selected and parameterized with respect to two randomly generated functions in the TRG to create two input signals, represented by curves in $\mathbb{R}^n$. For each input signal, $\tau^*(t)$ is computed from $\mathfrak{g}(t)$ and used to recover the UST parameterization of the sequence. The optimally reparameterized signals can them be compared element-wise using the error metric in~(\ref{error_seq}).}
\label{fig:demo_video_reparam}
\end{figure*}

We compared the performance of GORA with the DTW algorithm and implementations of the FastDTW algorithm with radii of 1, 5, and 20. The procedures with which we performed comparisons using both synthetic trajectories in $\mathbb{R}^3$ and video sequences are described as follows: For a given number of time instances, we randomly selected 50 different template signals. For each template signal, two initial parameterizations in the TRG were randomly generated and used to parameterize the original signal, creating 50 pairs of input signals, which were then fed to GORA and the DTW and FastDTW algorithms.

To ensure fair comparisons between algorithms, we use a modified version of GORA designed for the pairwise comparison of two signals, which is outlined in Fig. \ref{fig:demo_video_reparam}. This version accepts two input signals, $X_1(t)$ and $X_2(t)$, computes in parallel to their respective UST reparameterizations as defined in Algorithm \ref{algo1}, i.e. $X_1^*(t)$ and $X_2^*(t)$, and outputs the error between the two UST reparameterizations given by~(\ref{error_seq}). Similarly, we normalized the accumulated cost error output by the DTW and FastDTW algorithms under the Euclidean norm by dividing it by the length of the optimal warping path. Run time comparisons were performed using the clock module in Python's time package. Given two input signals, we defined the run time (what we called computational efficiency) to be the time it took each algorithm to output the error between them. 

\subsection{Results}

\begin{figure*}[!t] 
\centering
\begin{subfigure}{.48\linewidth}
\includegraphics[width=\textwidth]{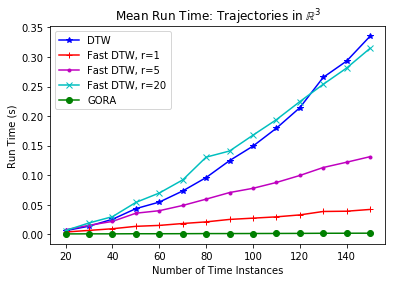}
\caption{Mean run time}
\label{fig:traj_run_time}
\end{subfigure}
\hskip.5em
\begin{subfigure}{.48\linewidth}
\includegraphics[width=\textwidth]{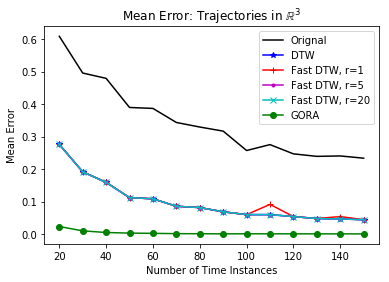}
\caption{Mean error}
\label{fig:traj_error}
\end{subfigure}
\vskip \baselineskip
\begin{subfigure}{.48\linewidth}
\includegraphics[width=\textwidth]{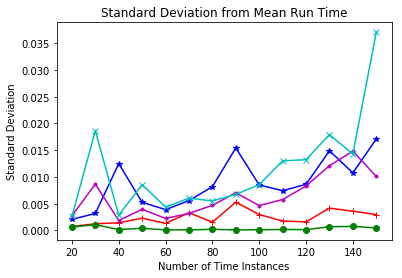}
\caption{Deviation from mean run time}
\label{fig:traj_run_time_std}
\end{subfigure}
\hskip.5em
\begin{subfigure}{.48\linewidth}
\includegraphics[width=\textwidth]{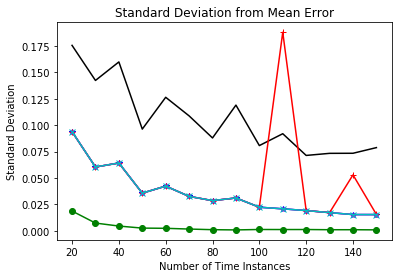}
\caption{Deviation from mean error}
\label{fig:traj_error_std}
\end{subfigure}

\caption{Algorithm performance: synthetic trajectories in $\mathbb{R}^3$.} \label{traj_comp}
\end{figure*}

\begin{figure*}[!t] 
\centering
\begin{subfigure}{.48\linewidth}
\includegraphics[width=\textwidth]{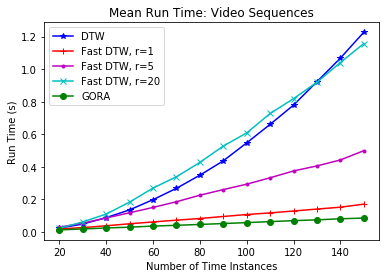}
\caption{Mean run time}
\label{fig:vid_run_time}
\end{subfigure}
\hskip.5em
\begin{subfigure}{.48\linewidth}
\includegraphics[width=\textwidth]{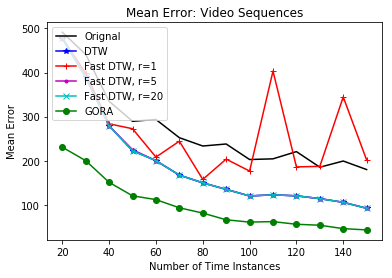}
\caption{Mean error}
\label{fig:vid_error}
\end{subfigure}
\vskip \baselineskip
\begin{subfigure}{.48\linewidth}
\includegraphics[width=\textwidth]{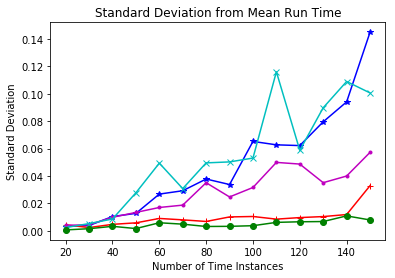}
\caption{Deviation from mean run time}
\label{fig:vid_run_time_std}
\end{subfigure}
\hskip.5em
\begin{subfigure}{.48\linewidth}
\includegraphics[width=\textwidth]{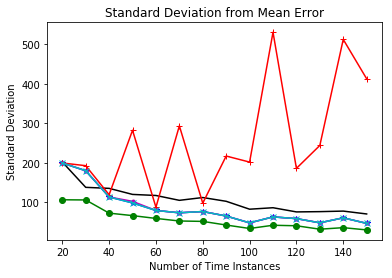}
\caption{Deviation from mean error}
\label{fig:vid_error_std}
\end{subfigure}

\caption{Algorithm performance: vectorized video sequences from the Weizmann Database.} \label{vid_comp}
\end{figure*}

Fig. \ref{traj_comp} and Fig. \ref{vid_comp} compare the performance of GORA and the DTW and FastDTW algorithms using signals in the form of trajectories in $\mathbb{R}^3$ and vectorized video sequences, respectively. Figs. \ref{fig:traj_run_time} and \ref{fig:vid_run_time} show the mean run time of each algorithm from 20 to 150 time instances. Figs. \ref{fig:traj_run_time_std} and \ref{fig:vid_run_time_std} show the corresponding standard deviations from the mean run times for each algorithm. 
	
With both trajectories in $\mathbb{R}^3$ and video sequences, as the total number of time instances increases,  DTW's run time grows quadratically (i.e. $O(T^2)$ complexity) while all iterations of the FastDTW algorithm and GORA achieve linear complexity (i.e. $O(T)$). However, in both cases GORA's run time is less than that of all the DTW methods, and GORA's complexity grows more slowly than the fastest implementation of FastDTW (radius $=$ 1). In addition, GORA's run time has a similar degree of stability (in the sense of smaller deviations from the overall mean run time) as that of the FastDTW implementation with radius $=$ 1, and remains significantly more stable than other DTW methods. 

Figs. \ref{fig:traj_error} and \ref{fig:vid_error} show the mean error  between signal pairs given by each algorithm from 20 to 150 time instances. Figs. \ref{fig:traj_error_std} and \ref{fig:vid_error_std} show the corresponding standard deviations from the mean error for each algorithm. In both cases, GORA is significantly more accurate (in the sense that the computed error between signal pairs known to represent the same dynamic phenomena is small) than the DTW algorithm and all implementations of the FastDTW algorithm. It was often the case that the DTW algorithm and the implementations of the FastDTW algorithm gave identical errors, since it is possible for the FastDTW algorithm to construct the same accumulated cost matrix as the DTW algorithm. 

The authors believe that the disparity in accuracy between GORA and the implementation of the FastDTW algorithm with radius $=$ 1 is especially significant. Since the error produced by the implementation of the FastDTW algorithm with radius $=$ 1 is both highly inaccurate and unstable (in the sense of large deviations from the mean error), especially for input signals in the form of video sequences, this suggests that an effective implementation of the FastDTW algorithm requires a larger radius. As such, the run time disparity between an effective FastDTW implementation and GORA is likely somewhere between the implementations of FastDTW with radius $=$ 1 and radius $=$ 5. 

That being said, these results constitute only an initial analysis with two types of elementary data. However, they do suggest that GORA has potential to be a highly effective framework for signal comparison and action recognition. 

\section{DISCUSSION}
A crucial difference between GORA and the DTW and FastDTW algorithms is GORA's reliance on interpolation to recover the UST reparameterization of the input signal. Depending on the context, this can be an advantage or disadvantage for the GORA framework. For example, consider the problem of signal comparison or action recognition over a space of signals where computing the error between signals at a given time instance is itself computationally expensive to perform. By interpolating, GORA only has to compute the pairwise error between signals no more than $T$ times, where $T$ is the total number of time instances. On the other hand, all DTW methods will have to compute this error between $O(T)$ and $O(T^2)$ times. If GORA's chosen method of interpolation is relatively inexpensive, this could give it a significant run time advantage over DTW methods. 

However, this could easily become a disadvantage for GORA if the chosen method of interpolation is relatively expensive compared to the computation the pairwise error between signals at a given time instance. In particular, this might serve to explain why GORA's run time advantage over the FastDTW implementation with radius $=$ 1 is smaller with video sequences than with trajectories in $\mathbb{R}^3$. For trajectories in $\mathbb{R}^3$,  GORA only performs three instances of linear interpolation --- one in each dimension of the trajectory. In contrast, for video sequences GORA performs $n = width \times height$ instances of linear interpolation along each dimension of the trajectory in $\mathbb{R}^{n}$ representing the vectorized video sequence. While this also means that DTW methods have to perform pairwise error computations with larger signals, it's likely that the cost due to increasing the instances of interpolation outweighs the cumulative cost of computing the euclidean norm in $\mathbb{R}^n$.  The videos in the Weizmann Database are relatively small ($180 \times 144$ pixels) and it is unknown with video sequences of larger dimensions whether GORA's mean run time would remain favorable relative to other methods. 

	The GORA framework is very much an active work-in-progress. Currently, we are exploring the GORA's potential in providing a foundation for a more robust algorithm able to minimize or eliminate nuisance parameters while simultaneously reparameterizing signals to a UST \cite{chirikjian2017signal}. The development of such an algorithm able to inherently compensate for perturbations such as noise or motion artifacts while maintaining a similar linear complexity to GORA would mark an important milestone toward the goal of robust robotic action recognition of human motions in real-time.  Additionally, a well-known strength of DTW methods is their ability to effectively compare signals with different numbers of time instances. This is something we have yet to consider in our implementations of GORA and a topic we plan to address in our future work.

\section{CONCLUSIONS}
In this paper, we introduced the  \textit{Globally Optimal Reparameterization Algorithm} (GORA) for signal alignment and comparison, based on a recently proposed novel mathematical framework \cite{chirikjian2017signal}. This algorithm reparameterizes signals to a \textit{universal standard timescale} (UST), allowing for element-wise comparisons between multiple signals at each instance of timewith linear time complexity of $O(T)$. In particular, we define procedures for applying this algorithm to characterize and compare signals in the form of real trajectories and video sequences.

Our experimental results have both provided a numerical validation of GORA's theoretical basis and suggested that the GORA framework has potential to become a viable alternative to DTW methods for signal comparison and action recognition purposes. In particular we showed that for signals in the form of real trajectories in $\mathbb{R}^3$ and vectorized video sequences with a fixed number of time instances, GORA's computational complexity is less than that of the FastDTW algorithm with radius$=$ 1 and that GORA's accuracy in matching signals representing fundamentally the same phenomena exceeds that of both the DTW algorithm and implementations of the FastDTW algorithm with radii of 1, 5, and 20.

\section{ACKNOWLEDGMENTS}
The authors would like to thank Dr. Jin Seob Kim and Ms. Mengdi Xu for useful discussions that contributed to this work. This work was performed under National Science Foundation grant IIS-1619050 and Office of Naval Research Award N00014-17-1-2142.

\bibliography{reference}

\begin{thebibliography}{10}
\providecommand{\url}[1]{#1}
\csname url@samestyle\endcsname
\providecommand{\newblock}{\relax}
\providecommand{\bibinfo}[2]{#2}
\providecommand{\BIBentrySTDinterwordspacing}{\spaceskip=0pt\relax}
\providecommand{\BIBentryALTinterwordstretchfactor}{4}
\providecommand{\BIBentryALTinterwordspacing}{\spaceskip=\fontdimen2\font plus
\BIBentryALTinterwordstretchfactor\fontdimen3\font minus
  \fontdimen4\font\relax}
\providecommand{\BIBforeignlanguage}[2]{{%
\expandafter\ifx\csname l@#1\endcsname\relax
\typeout{** WARNING: IEEEtran.bst: No hyphenation pattern has been}%
\typeout{** loaded for the language `#1'. Using the pattern for}%
\typeout{** the default language instead.}%
\else
\language=\csname l@#1\endcsname
\fi
#2}}
\providecommand{\BIBdecl}{\relax}
\BIBdecl

\bibitem{sakoe1978dynamic}
H.~Sakoe and S.~Chiba, ``Dynamic programming algorithm optimization for spoken
  word recognition,'' \emph{IEEE Transactions on Acoustics, Speech, and Signal
  Processing}, vol.~26, no.~1, pp. 43--49, Feb 1978.

\bibitem{shaikh2017recognition}
H.~Shaikh, L.~C. Mesquita, S.~D. C.~S. Araujo, P.~Student, and A.~P. Professor,
  ``Recognition of isolated spoken words and numeric using mfcc and dtw,''
  \emph{International Journal of Engineering Science}, vol. 10539, 2017.

\bibitem{sempena2011human}
S.~Sempena, N.~U. Maulidevi, and P.~R. Aryan, ``Human action recognition using
  dynamic time warping,'' in \emph{Proceedings of the 2011 International
  Conference on Electrical Engineering and Informatics}, July 2011, pp. 1--5.

\bibitem{keogh2000scaling}
E.~J. Keogh and M.~J. Pazzani, ``Scaling up dynamic time warping to massive
  datasets,'' in \emph{Principles of Data Mining and Knowledge Discovery},
  J.~M. {\.{Z}}ytkow and J.~Rauch, Eds.\hskip 1em plus 0.5em minus 0.4em\relax
  Berlin, Heidelberg: Springer Berlin Heidelberg, 1999, pp. 1--11.

\bibitem{zhu2017robust}
\BIBentryALTinterwordspacing
T.~Zhu, Q.~Zhao, W.~Wan, and Z.~Xia, ``Robust regression-based motion
  perception for online imitation on humanoid robot,'' \emph{International
  Journal of Social Robotics}, vol.~9, no.~5, pp. 705--725, Nov 2017. [Online].
  Available: \url{https://doi.org/10.1007/s12369-017-0416-9}
\BIBentrySTDinterwordspacing

\bibitem{keogh2001derivative}
E.~J. Keogh and M.~J. Pazzani, ``Derivative dynamic time warping,'' in
  \emph{Proceedings of the 2001 SIAM International Conference on Data
  Mining}.\hskip 1em plus 0.5em minus 0.4em\relax SIAM, 2001, pp. 1--11.

\bibitem{salvador2007toward}
S.~Salvador and P.~Chan, ``Fastdtw: Toward accurate dynamic time warping in
  linear time and space,'' \emph{Intelligent Data Analysis}, vol.~11, no.~5,
  pp. 561--580, 2007.

\bibitem{keogh2005exact}
\BIBentryALTinterwordspacing
E.~Keogh and C.~A. Ratanamahatana, ``Exact indexing of dynamic time warping,''
  \emph{Knowledge and Information Systems}, vol.~7, no.~3, pp. 358--386, Mar
  2005. [Online]. Available: \url{https://doi.org/10.1007/s10115-004-0154-9}
\BIBentrySTDinterwordspacing

\bibitem{zhou2012generalized}
F.~Zhou and F.~D. la~Torre, ``Generalized time warping for multi-modal
  alignment of human motion,'' in \emph{2012 IEEE Conference on Computer Vision
  and Pattern Recognition}, June 2012, pp. 1282--1289.

\bibitem{chirikjian2017signal}
G.~S. Chirikjian, ``Signal classification in quotient spaces via globally
  optimal variational calculus,'' in \emph{Computer Vision and Pattern
  Recognition Workshops (CVPRW), 2017 IEEE Conference on}.\hskip 1em plus 0.5em
  minus 0.4em\relax IEEE, 2017, pp. 735--743.

\bibitem{ActionsAsSpaceTimeShapes_iccv05}
M.~Blank, L.~Gorelick, E.~Shechtman, M.~Irani, and R.~Basri, ``Actions as
  space-time shapes,'' in \emph{The Tenth IEEE International Conference on
  Computer Vision (ICCV'05)}, 2005, pp. 1395--1402.

\bibitem{ActionsAsSpaceTimeShapes_pami07}
L.~Gorelick, M.~Blank, E.~Shechtman, M.~Irani, and R.~Basri, ``Actions as
  space-time shapes,'' \emph{Transactions on Pattern Analysis and Machine
  Intelligence}, vol.~29, no.~12, pp. 2247--2253, December 2007.

\end{thebibliography}
\bibliographystyle{IEEEtran}

\end{document}